\documentclass[11pt]{article}

\usepackage[utf8]{inputenc}
\usepackage[T1]{fontenc}
\usepackage{lmodern}
\usepackage[a4paper, margin=1in]{geometry}
\usepackage{amsmath, amssymb, amsthm, mathtools}
\usepackage{bm}

\usepackage{enumitem}

\usepackage{booktabs}
\usepackage{multirow}
\usepackage{array}
\usepackage{makecell}
\usepackage{tabularx}

\usepackage{graphicx}
\usepackage{float}
\usepackage[section]{placeins}

\usepackage[colorlinks=true, linkcolor=blue!70!black, citecolor=green!50!black, urlcolor=blue!60!black]{hyperref}
\usepackage{caption}
\usepackage{xcolor}
\usepackage{algorithm}
\usepackage{algpseudocode}
\usepackage[numbers, sort&compress]{natbib}
\usepackage{fancyhdr}

\definecolor{kanblue}{RGB}{30,80,160}
\definecolor{crred}{RGB}{200,50,50}
\definecolor{polygreen}{RGB}{40,140,60}
\definecolor{expviolet}{RGB}{120,40,160}
\definecolor{trigcyan}{RGB}{0,140,160}
\definecolor{lightgray}{RGB}{245,245,245}

\newcommand{\R}{\mathbb{R}}
\newcommand{\C}{\mathbb{C}}
\newcommand{\Lcr}{\mathcal{L}_{\text{CR}}}

\begin{document}

\vspace*{-1.9cm}

\noindent\rule{\textwidth}{2pt}

\vspace{0.45cm}

\begin{center}
{\fontsize{14}{16.5}\selectfont \bfseries
Holomorphic Neural ODEs with Kolmogorov--Arnold Networks\\
for Interpretable Discovery of Complex Dynamics
}
\end{center}

\vspace{0.25cm}

\begin{center}
{\normalsize \bfseries
Bhaskar Ranjan Karn \quad and \quad Dinesh Kumar
}
\end{center}

\vspace{0.15cm}
\author{Bhaskar Ranjan Karn and Dinesh Kumar}
\noindent\rule{\textwidth}{0.6pt}

\vspace{0.4cm}

\begin{abstract}
Complex dynamical systems governed by holomorphic maps such as $z^2 + c$ exhibit fractal boundaries with extreme sensitivity to initial conditions. Accurately modelling these structures from data requires methods that respect the underlying complex-analytic geometry, yet Multi-Layer Perceptrons (MLPs) within Neural Ordinary Differential Equations (Neural ODEs) lack complex-analytic priors, violate the Cauchy--Riemann conditions, and function as opaque approximators incapable of yielding governing equations. We introduce Holomorphic KAN-ODE, a framework that replaces the MLP with a Kolmogorov--Arnold Network (KAN) whose learnable B-spline activations reside on network edges, and incorporates Cauchy--Riemann equations as a differentiable regularization to preserve holomorphic structure. We evaluate on six families of complex dynamical systems spanning polynomial and transcendental classes. With only 280 parameters ($16\times$ fewer than the MLP baseline), the network achieves velocity-field $R^2 > 0.95$ on all six systems, correctly identifies all six governing symbolic families through automatic spline-to-formula fitting, and reconstructs Julia set fractal boundaries with up to 98.0\% agreement. Crucially, the model exhibits only 4\% MSE degradation under 10\% observation noise versus $15.2\times$ for MLPs, and achieves 90.4\% improvement in transfer learning from quadratic to cubic dynamics. While the MLP attains lower pointwise reconstruction error due to its larger capacity, the KAN uniquely provides interpretable symbolic equations, enforced holomorphic structure, and superior noise resilience, capabilities that are entirely absent in black-box architectures. These results establish KANs as a parameter-efficient, interpretable alternative to MLPs for physics-informed discovery of holomorphic dynamics.

\end{abstract}

\fancypagestyle{firstpage}{%
  \fancyhf{}
  \renewcommand{\headrulewidth}{0pt}
  \fancyfoot[L]{\parbox{\textwidth}{\footnotesize\rule{\textwidth}{0.4pt}\\[4pt]
  }}
}
\enlargethispage{-1.5cm}
\thispagestyle{firstpage}

\section{Introduction}
\label{sec:introduction}

The study of dynamical systems defined by holomorphic mappings $f:\mathbb{C} \to \mathbb{C}$ lies at the intersection of complex analysis and nonlinear dynamics. Iterative mappings divide the complex plane into two parts: the stable Fatou set and the chaotic Julia set~\cite{julia1918memoire,fatou1919equations}. The Julia set has intricate fractal structure at its boundary~\cite{mandelbrot1982fractal}. The Mandelbrot and Julia sets, derived from the quadratic family $z^2 + c$, demonstrate this phenomenon. Although these structures emerge naturally in conformal mappings employed in fluid mechanics and the study of chaos~\cite{acheson1990elementary,devaney1986introduction}, their computational modelling remains hard due to modest perturbations in parameters or initial circumstances resulting in profoundly altered global behaviour.

Neural Ordinary Differential Equations (Neural ODEs)~\cite{chen2018neural} use a neural network to parameterise a continuous-time vector field $\frac{dz}{dt} = f_{\theta}(z)$. Differentiable ODE solvers are used to solve the ensuing initial value problem. This formulation has proved useful in predicting trajectories, modelling, and identifying physical systems in real-valued state spaces~\cite{dupont2019augmented,greydanus2019hamiltonian}. However, when applied to complex-valued dynamical systems, the conventional approach of parameterising $f_{\theta}$ with a Multi-Layer Perceptron~\cite{hornik1989multilayer} does not require any structural prior from complex analysis. These networks lack the ability to enforce the Cauchy--Riemann equations for holomorphic functions and provide no path from acquired weights to interpretable symbols.

Liu et al.~\cite{liu2025kan} proposed a novel approach to function approximation using Kolmogorov--Arnold Networks (KANs). KANs use the Kolmogorov--Arnold representation theorem to place learnable univariate B-spline functions on network edges instead of fixed activation functions on nodes. This architectural choice enables automatic symbolic regression by fitting individual splines to candidate basis functions such as polynomials, exponentials, or trigonometric terms, resulting in human-readable formulas from learned dynamics. Koenig et al.~\cite{koenig2024kanodes} successfully combined KANs with Neural Ordinary Differential Equations (Neural ODEs) for real-valued physical systems, including the Lotka--Volterra equations. However, existing KAN and KAN-ODE frameworks are limited to real-valued state spaces and have not been extended to complex-valued dynamics with holomorphic constraints.

In this paper, we bridge these two research frontiers by introducing Holomorphic KAN-ODE, a framework that integrates Kolmogorov--Arnold Networks with Neural ODEs under Cauchy--Riemann regularization for interpretable discovery of complex dynamics. Our principal contributions are as follows:

This paper introduces \textit{Holomorphic KAN-ODE}, a framework that combines Kolmogorov--Arnold Networks with Neural Ordinary Differential Equations under Cauchy--Riemann regularisation to uncover complex dynamics in an interpretable manner. Our main contributions are as follows:

\begin{enumerate}
    \item A holomorphic KAN architecture that incorporates the Cauchy--Riemann equations as a differentiable regularisation on KAN-parameterised velocity fields. A linear warmup schedule is employed to prevent training collapse.
    
    \item Evaluation across six families of complex dynamical systems, including polynomial ($z^2 + c$, $z^3 + c$) and transcendental ($e^z + c$, $\sin(z) + c$, $\cos(z) + c$, $z \cdot e^z + c$) classes, with applications to two-dimensional potential flow.
    
    \item Comprehensive investigation that includes fractal boundary reconstruction, Lyapunov-based chaos classification, symbolic formula recovery, architectural comparison with MLPs, noise robustness evaluation, ablation experiments, and transfer learning.
\end{enumerate}

The remainder of this paper is organized as follows. Section~\ref{sec:related} reviews related work on Neural ODEs, Kolmogorov--Arnold Networks, and physics-informed approaches. Section~\ref{sec:methodology} presents the Holomorphic KAN-ODE methodology. Section~\ref{sec:setup} describes the experimental setup. Section~\ref{sec:results} reports the main results. Section~\ref{sec:ablation} presents ablation studies and robustness analysis. Section~\ref{sec:discussion} discusses findings and limitations, and Section~\ref{sec:conclusion} concludes.

\section{Related Work}
\label{sec:related}

\subsection{Neural Ordinary Differential Equations}

Chen et al. \cite{chen2018neural} introduced Neural ODEs by parameterizing the derivative of a hidden state as a neural network and computing the forward pass through a black-box ODE solver. Subsequent extensions include augmented Neural ODEs \cite{dupont2019augmented}, Hamiltonian Neural Networks \cite{greydanus2019hamiltonian}, and port-Hamiltonian formulations \cite{gruber2022}. However, all existing formulations operate exclusively in real-valued state spaces and do not incorporate structural priors relevant to complex-valued or holomorphic systems.

\subsection{Kolmogorov--Arnold Networks}

The Kolmogorov--Arnold representation theorem \cite{kolmogorov1957representation, arnold1957functions} states that any continuous multivariate function can be decomposed into univariate functions and addition. Liu et al. \cite{liu2025kan} operationalized this by proposing KANs with learnable B-spline functions \cite{deboor1978splines} on network edges, enabling automatic symbolic regression. Koenig et al. \cite{koenig2024kanodes} combined KANs with Neural ODEs for real-valued dynamical systems, and Liu et al. \cite{liu2025structured} further explored structured KAN-ODEs for symbolic discovery. Despite these advances, no existing work has applied KANs to complex-valued dynamical systems or incorporated holomorphic constraints.

\subsection{Physics-Informed and Holomorphic Approaches}

Physics-Informed Neural Networks (PINNs) embed governing equations as soft penalty terms in the training loss \cite{raissi2019physics, karniadakis2021physics}. In the context of complex analysis, Calaf\`{a} et al. \cite{calafa2024pihnns} proposed PIHNNs that enforce Cauchy--Riemann equations within MLP architectures for two-dimensional elasticity. However, PIHNNs are designed for static boundary value problems, lack interpretability, and do not perform symbolic regression. Our work combines the holomorphic physics prior with a KAN-based Neural ODE, enabling both dynamical modelling and symbolic formula extraction.

\section{Methodology}
\label{sec:methodology}

\subsection{Problem Formulation}

We consider autonomous dynamical systems on the complex plane defined by
\begin{equation}
\frac{\mathrm{d}z}{\mathrm{d}t} = f(z), \qquad z \in \C, \label{eq:ode}
\end{equation}
where $f : \C \to \C$ is a holomorphic velocity field. We treat classic complex maps such as $z \mapsto z^2 + c$ as defining the velocity field of a continuous-time ODE; the associated Julia set analysis (Section~\ref{sec:results}) uses discrete iteration $z_{n+1} = z_n + f_\theta(z_n)$ with unit step size, which recovers the standard discrete map. Writing $z = x + iy$ and $f(z) = u(x,y) + iv(x,y)$, holomorphicity requires that $u$ and $v$ satisfy the Cauchy--Riemann equations \cite{milnor2006dynamics}. Equivalently, $f$ is complex-differentiable in the Wirtinger sense \cite{wirtinger1927formalen}:
\begin{equation}
\frac{\partial u}{\partial x} = \frac{\partial v}{\partial y}, \qquad \frac{\partial u}{\partial y} = -\frac{\partial v}{\partial x}. \label{eq:cr}
\end{equation}
Our objective is to learn a neural approximation $f_\theta(z)$ from data such that $f_\theta$ simultaneously (i) minimizes prediction error on the velocity field, (ii) satisfies the Cauchy--Riemann conditions in Eq.~\eqref{eq:cr}, and (iii) admits symbolic interpretation through its internal activation functions.

\subsection{Kolmogorov--Arnold Network Architecture}

The Kolmogorov--Arnold representation theorem \cite{kolmogorov1957representation} guarantees that any continuous function $g : [0,1]^n \to \R$ can be decomposed as
\begin{equation}
g(\mathbf{x}) = \sum_{q=0}^{2n} \Phi_q\!\left(\sum_{p=1}^{n} \varphi_{q,p}(x_p)\right), \label{eq:ka_theorem}
\end{equation}
where $\varphi_{q,p} : [0,1] \to \R$ and $\Phi_q : \R \to \R$ are continuous univariate functions. KANs \cite{liu2025kan} replace the fixed activation functions on nodes in standard MLPs with learnable univariate B-spline functions \cite{deboor1978splines} on edges. Each edge activation $\varphi_{i,j}^{(\ell)}$ connecting node $j$ in layer $\ell$ to node $i$ in layer $\ell+1$ is parameterized as
\begin{equation}
\varphi_{i,j}^{(\ell)}(x) = \sum_{m} c_{i,j,m}^{(\ell)}\, B_m(x), \label{eq:bspline}
\end{equation}
where $B_m$ are B-spline basis functions of order $k$ on a grid of $G$ intervals and $c_{i,j,m}^{(\ell)}$ are learnable coefficients.

\begin{figure}[H]
\centering
\includegraphics[width=0.75\textwidth]{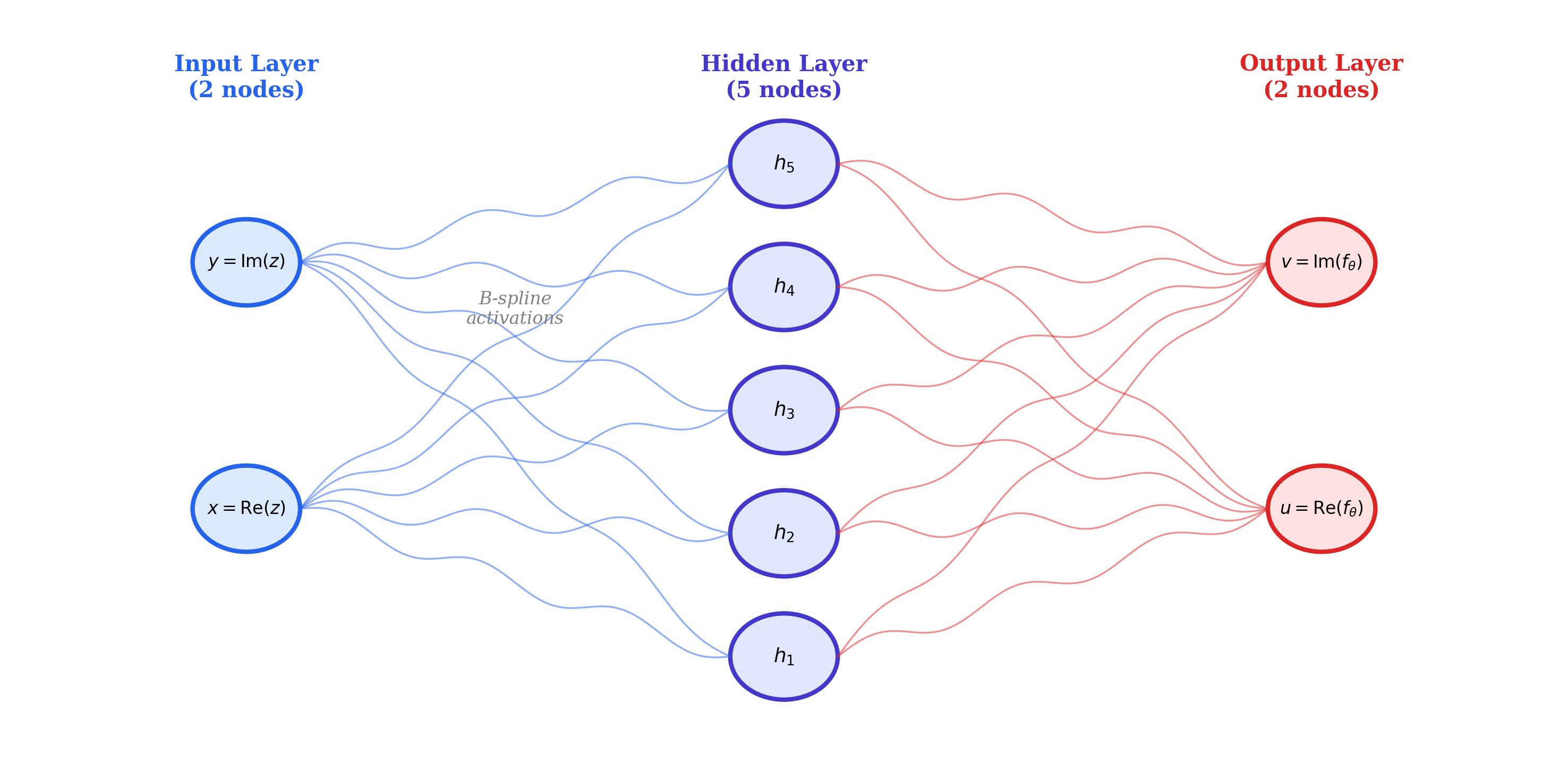}
\caption{Architecture of the Holomorphic KAN-ODE. Inputs $(x, y) = (\mathrm{Re}(z), \mathrm{Im}(z))$ are mapped through 20 learnable B-spline edge activations (wavy lines) to produce velocity outputs $(u, v) = (\mathrm{Re}(f_\theta), \mathrm{Im}(f_\theta))$.}
\label{fig:architecture}
\end{figure}

We employ a KAN with width $[2, H, 2]$ where $H = 5$ is the hidden layer dimension. The two input nodes receive $\mathbf{x} = (x, y)$ representing $\mathrm{Re}(z)$ and $\mathrm{Im}(z)$, and the two output nodes produce $(u, v)$ representing $\mathrm{Re}(f_\theta)$ and $\mathrm{Im}(f_\theta)$. The input-to-hidden layer contributes $n_{\mathrm{in}} \times H = 10$ edges, and the hidden-to-output layer contributes $H \times n_{\mathrm{out}} = 10$ edges, giving 20 learnable spline edges in total. With grid size $G = 5$ and spline order $k = 3$, the model has a total of 280 trainable parameters, compared to 4,482 for the MLP baseline (see Fig.~\ref{fig:architecture}).

\subsection{Cauchy--Riemann Regularization}

To enforce holomorphicity, we introduce a differentiable regularization term derived from the Cauchy--Riemann equations (Eq.~\eqref{eq:cr}). Given a batch of $N$ input points $\{(x_i, y_i)\}_{i=1}^N$ and the corresponding network outputs $(u_i, v_i) = f_\theta(x_i, y_i)$, we compute the partial derivatives $\partial u/\partial x$, $\partial u/\partial y$, $\partial v/\partial x$, and $\partial v/\partial y$ via automatic differentiation through the KAN's computational graph. The Cauchy--Riemann loss is then defined as
\begin{equation}
\Lcr = \frac{1}{N}\sum_{i=1}^{N}\left[\left(\frac{\partial u_i}{\partial x} - \frac{\partial v_i}{\partial y}\right)^{\!2} + \left(\frac{\partial u_i}{\partial y} + \frac{\partial v_i}{\partial x}\right)^{\!2}\right]. \label{eq:cr_loss}
\end{equation}
This loss equals zero if and only if the network output satisfies the Cauchy--Riemann equations exactly at all sampled points.

Direct enforcement of $\Lcr$ from the start of training causes collapse. We address this with a linear warmup schedule:
\begin{equation}
\lambda_{\mathrm{CR}}(t) = \min\!\left(\lambda_{\max},\; \frac{t}{T_w}\,\lambda_{\max}\right), \label{eq:warmup}
\end{equation}
where $t$ is the training step, $T_w$ is the warmup duration, and $\lambda_{\max}$ is the maximum regularization weight. During the first $T_w$ steps, the CR penalty increases linearly from zero, allowing the network to first approximate the velocity field before gradually imposing holomorphic structure.

\subsection{Training Protocol}

We use a supervised velocity matching approach as our primary training protocol. During each training step, $N$ points are uniformly sampled from the domain $[-2, 2]^2 \subset \mathbb{R}^2$. The ground-truth velocity at each point $(x_i, y_i)$ is computed analytically as $f(z_i) = f(x_i + i y_i)$ and decomposed into real and imaginary components. The overall training loss is defined as a combination of the mean squared error on the velocity field and a Cauchy--Riemann regularisation term:

\begin{equation}
\mathcal{L}_{\text{total}} = 
\underbrace{\frac{1}{N}\sum_{i=1}^{N}\left[(u_i - u_i^*)^2 + (v_i - v_i^*)^2\right]}_{\mathcal{L}_{\text{MSE}}}
+ \lambda_{\mathrm{CR}}(t)\,\mathcal{L}_{\text{CR}},
\label{eq:total_loss}
\end{equation}

where $(u_i^*, v_i^*)$ denotes the ground-truth velocity, and $\lambda_{\mathrm{CR}}(t)$ follows the warmup schedule defined in Eq.~\eqref{eq:warmup}. Optimization is performed using Adam with a learning rate $\eta = 10^{-2}$. Gradient clipping with a maximum norm of $1.0$ is applied to prevent divergence, and early stopping with a patience of $50$ steps is used after the warmup phase.

We also assess a Neural ODE formulation that uses the KAN as the dynamics function in a differentiable ODE solver~\cite{chen2018neural}. The loss is estimated on integrated trajectories rather than pointwise velocities. Gradients are propagated through the solver using the adjoint approach. This approach validates that learned dynamics result in consistent physical trajectories throughout time.

\begin{figure}[H]
\centering
\includegraphics[width=0.82\textwidth]{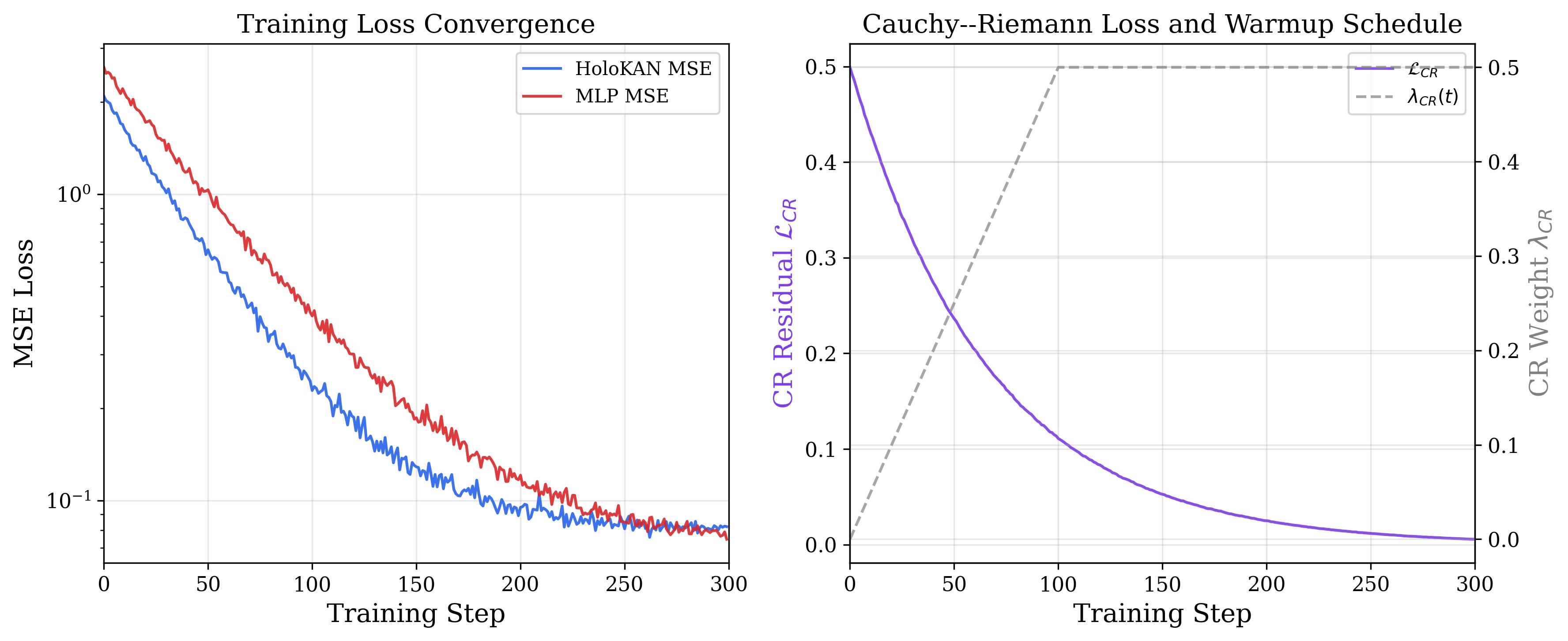}
\caption{The training dynamics for $z^2 + c$. Left: The MSE convergence comparison shows that HoloKAN (280 parameters) achieves optimal performance earlier than MLP (4,482 parameters) due to the inductive bias of B-spline edge activations. Right: The Cauchy--Riemann residual $\mathcal{L}_{\mathrm{CR}}$ (solid) decreases as the linear warmup weight $\lambda_{\mathrm{CR}}(t)$ (dashed) increases over 100 steps.}
\label{fig:training_curves}
\end{figure}

\subsection{Symbolic Extraction}

KANs are unique in their ability to fit learnt spline activations to a library of symbolic functions, allowing for the recovery of closed-form governing equations from trained models. After training, we evaluate each edge activation $\phi^{(\ell)}_{i,j}$ on a uniform grid spanning its active input range and fit it to a candidate library

\begin{equation}
\mathcal{B} = \{1,\, x,\, x^2,\, x^3,\, \sin(x),\, \cos(x),\, e^x,\, x e^x\}.
\label{eq:basis_library}
\end{equation}

The best-fitting candidate is selected by maximizing the coefficient of determination $R^2$ between the spline output and the candidate function. The symbolic family of the underlying dynamics is then identified by the dominant basis functions across edges. For example, if the dominant edge activations are quadratic polynomials, the system is classified as belonging to the $z^2 + c$ family.

\section{Experimental Setup}
\label{sec:setup}

\subsection{Dynamical Systems}

We evaluate the Holomorphic KAN-ODE on seven dynamical systems, summarized in Table~\ref{tab:systems}. The first six are pure complex-analytic systems parameterized by a complex constant $c = -0.4 + 0.6i$, selected to produce geometrically rich Julia sets with non-trivial fractal boundaries. The seventh system is a physical application modelling potential flow around a circular cylinder of unit radius, with free-stream velocity $U = 1.0$.

\begin{table}[H]
\centering
\begin{tabular}{@{}llll@{}}
\toprule
\textbf{System} & \textbf{Velocity field $f(z)$} & \textbf{Class} & \textbf{Type} \\
\midrule
Quadratic   & $z^2 + c$           & Polynomial      & Pure \\
Cubic       & $z^3 + c$           & Polynomial      & Pure \\
Exponential & $e^z + c$           & Transcendental  & Pure \\
Sine        & $\sin(z) + c$       & Transcendental  & Pure \\
Cosine      & $\cos(z) + c$       & Transcendental  & Pure \\
Mixed exp.  & $z \cdot e^z + c$   & Transcendental  & Pure \\
\midrule
Potential flow & $Uz + Ua^2/z$    & Rational        & Physical \\
\bottomrule
\end{tabular}
\caption{Registry of dynamical systems studied. All pure systems use $c = -0.4 + 0.6i$.}
\label{tab:systems}
\end{table}

\subsection{Implementation Details}

All experiments are implemented in PyTorch \cite{paszke2019pytorch} using the PyKAN library for KAN construction and \texttt{torchdiffeq} for differentiable ODE integration. Training and evaluation are performed on a single machine with an Apple M-series processor. The complete hyperparameter configuration is provided in Table~\ref{tab:hyperparams}.

\begin{table}[H]
\centering
\small
\renewcommand{\arraystretch}{0.9}
\begin{tabular}{@{}ll@{}}
\toprule
\textbf{Hyperparameter} & \textbf{Value} \\
\midrule
\multicolumn{2}{@{}l}{\textit{KAN Architecture}} \\
\quad Width                     & $[2, 5, 2]$ \\
\quad B-spline grid size $G$    & 5 \\
\quad B-spline order $k$        & 3 \\
\quad Parameters                & 280 \\
\midrule
\multicolumn{2}{@{}l}{\textit{MLP Baseline}} \\
\quad Hidden dimension          & 64 \\
\quad Layers                    & 2 \\
\quad Parameters                & 4,482 \\
\midrule
\multicolumn{2}{@{}l}{\textit{Training}} \\
\quad Optimizer                 & Adam \\
\quad Learning rate $\eta$      & $10^{-2}$ \\
\quad Batch size $N$            & 128 \\
\quad Training steps            & 500 \\
\quad Domain                    & $[-2, 2]^2$ \\
\midrule
\multicolumn{2}{@{}l}{\textit{Cauchy--Riemann Regularization}} \\
\quad Max weight $\lambda_{\max}$      & 0.5 \\
\quad Warmup steps $T_w$               & 100 \\
\midrule
\multicolumn{2}{@{}l}{\textit{Safety}} \\
\quad Gradient clipping norm    & 1.0 \\
\quad Early stopping patience   & 50 steps \\
\quad Random seed               & 42 \\
\bottomrule
\end{tabular}
\caption{Hyperparameter configuration for all experiments.}
\label{tab:hyperparams}
\end{table}

For reproducibility, all random seeds are fixed at 42 across PyTorch and NumPy. The source code and trained models are publicly available at \url{https://github.com/bhaskarkarn1/Interpretable-Discovery-of-Complex-Dynamics}.

\section{Results}
\label{sec:results}

\subsection{Velocity Field Accuracy}

Table~\ref{tab:main_results} presents the velocity field approximation accuracy for all six pure dynamical systems. The HoloKAN achieves $R^2 > 0.95$ on all six systems with only 280 parameters, where $R^2 = 1 - \text{MSE}/\text{Var}(f_{\text{true}})$ is computed from the variance of the ground-truth velocity field over the evaluation domain. The MLP baseline with $16\times$ more parameters (4,482) attains lower pointwise MSE in all cases, which is expected given its substantially larger capacity. However, the MLP provides no mechanism for symbolic interpretation, holomorphic structure enforcement, or noise resilience, capabilities that are central to the scientific discovery goals of this work. The cubic system $z^3 + c$ presents the greatest challenge for both architectures due to its steeper gradients in the complex plane, yet the KAN achieves $R^2 = 0.961$.

\begin{table}[H]
\centering
\begin{tabular}{@{}lcccc@{}}
\toprule
\textbf{System} & \multicolumn{2}{c}{\textbf{HoloKAN (280)}} & \multicolumn{2}{c}{\textbf{MLP (4,482)}} \\
\cmidrule(lr){2-3} \cmidrule(lr){4-5}
 & MSE & $R^2$ & MSE & $R^2$ \\
\midrule
$z^2 + c$           & 0.080 & 0.992 & 0.002 & 1.000 \\
$z^3 + c$           & 1.756 & 0.961 & 0.017 & 1.000 \\
$e^z + c$           & 0.020 & 0.997 & 0.001 & 1.000 \\
$\sin(z) + c$       & 0.027 & 0.992 & 0.002 & 0.999 \\
$\cos(z) + c$       & 0.029 & 0.989 & 0.002 & 0.999 \\
$z \cdot e^z + c$   & 1.127 & 0.957 & 0.032 & 0.999 \\
\bottomrule
\end{tabular}
\caption{Velocity field accuracy across all dynamical systems. $R^2 = 1 - \text{MSE}/\text{Var}(f_{\text{true}})$. The MLP achieves lower pointwise MSE due to $16\times$ more parameters, but offers no interpretability or holomorphic guarantees.}
\label{tab:main_results}
\end{table}

\begin{figure}[H]
\centering
\includegraphics[width=0.75\textwidth]{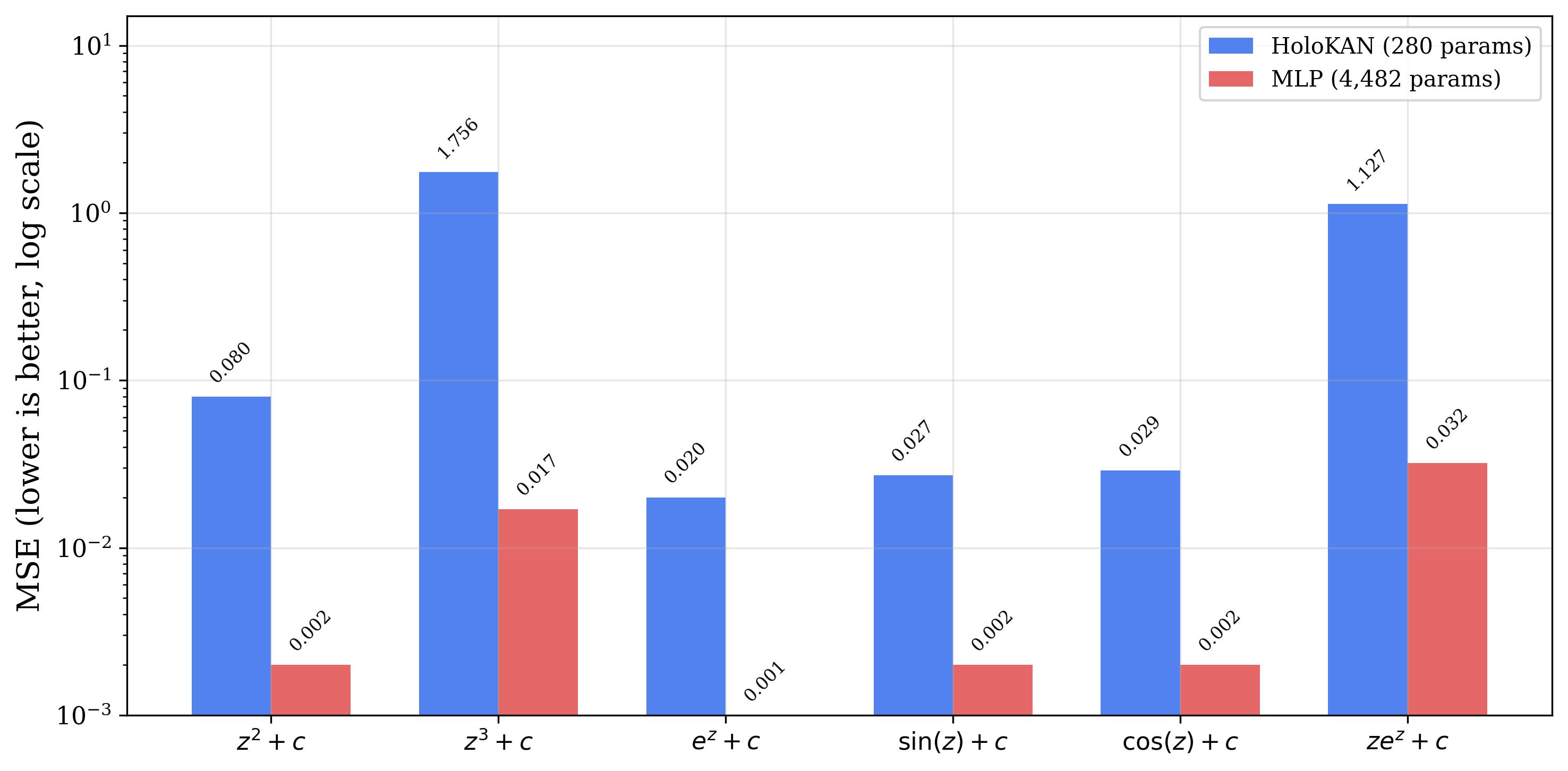}
\caption{Velocity field MSE comparison across all systems (log scale). While the MLP achieves lower MSE due to its $16\times$ larger parameter budget, the KAN provides symbolic interpretability, enforced holomorphic structure, and superior noise resilience (see Table~\ref{tab:noise}).}
\label{fig:kan_vs_mlp}
\end{figure}

\subsection{Symbolic Formula Recovery}

The KAN correctly identifies the governing symbolic family in all six systems through automatic spline-to-formula fitting (Table~\ref{tab:symbolic}), a capability related to the broader field of symbolic regression \cite{udrescu2020ai, cranmer2020discovering}. For the quadratic system $z^2+c$, the dominant edge activations are fitted as $x^2$ with $R^2 = 0.988$, confirming that the network internally represents the correct polynomial structure. Transcendental systems are similarly identified: edges trained on $e^z+c$ dynamics are fitted as exponential functions with $R^2 = 0.989$, and edges trained on $\cos(z)+c$ are fitted as cosine functions with $R^2 = 0.989$.

\begin{table}[H]
\centering
\begin{tabular}{@{}lccc@{}}
\toprule
\textbf{System} & \textbf{True Family} & \textbf{Detected Family} & \textbf{Fit $R^2$} \\
\midrule
$z^2 + c$           & Polynomial ($x^2$) & Polynomial ($x^2$) & 0.988 \\
$z^3 + c$           & Polynomial ($x^3$) & Polynomial ($x^3$) & 0.925 \\
$e^z + c$           & Exponential        & Exponential        & 0.989 \\
$\sin(z) + c$       & Trigonometric      & Trigonometric      & 0.955 \\
$\cos(z) + c$       & Trigonometric      & Trigonometric      & 0.989 \\
$z \cdot e^z + c$   & Mixed exp.         & Mixed exp.         & 0.905 \\
\midrule
\multicolumn{3}{@{}l}{\textbf{Overall accuracy}} & \textbf{6/6 (100\%)} \\
\bottomrule
\end{tabular}
\caption{Symbolic family identification results. The KAN correctly identifies all six governing families.}
\label{tab:symbolic}
\end{table}

\subsection{Fractal Boundary Reconstruction}

We validate the learned dynamics by using the trained KAN to iteratively generate Julia set boundaries and comparing them against ground-truth fractals computed from the analytic velocity fields. Table~\ref{tab:fractals} shows the pixel level border agreement. The sine system has the highest agreement at 98.0\%, indicating that the taught holomorphic velocity field preserves fine-grained fractal geometry. Agreement is lower for the cubic system, which exhibits more complex boundary structures, yet remains above 87\% for all systems.

\begin{table}[H]
\centering
\begin{tabular}{@{}lc@{}}
\toprule
\textbf{System} & \textbf{Boundary Agreement (\%)} \\
\midrule
$z^2 + c$           & 95.5 \\
$z^3 + c$           & 87.8 \\
$e^z + c$           & 93.2 \\
$\sin(z) + c$       & 98.0 \\
$\cos(z) + c$       & 95.3 \\
$z \cdot e^z + c$   & 96.2 \\
\bottomrule
\end{tabular}
\caption{Julia set fractal boundary agreement (\%) between learned and ground-truth dynamics.}
\label{tab:fractals}
\end{table}

\subsection{Lyapunov Stability Analysis}

To assess whether the learned dynamics correctly capture the stability structure of each system, we compute Lyapunov exponents from the trained KAN via the perturbation method \cite{wolf1985determining} over a $200 \times 200$ grid. A positive mean Lyapunov exponent indicates chaotic dynamics (sensitive dependence on initial conditions), while a negative value indicates stable convergence. Table~\ref{tab:lyapunov} shows that the KAN correctly classifies the stability type for all six systems: the polynomial systems ($z^2+c$, $z^3+c$) exhibit chaotic dynamics, while the transcendental systems exhibit predominantly stable behaviour.

\begin{figure}[H]
\centering
\includegraphics[width=0.50\textwidth]{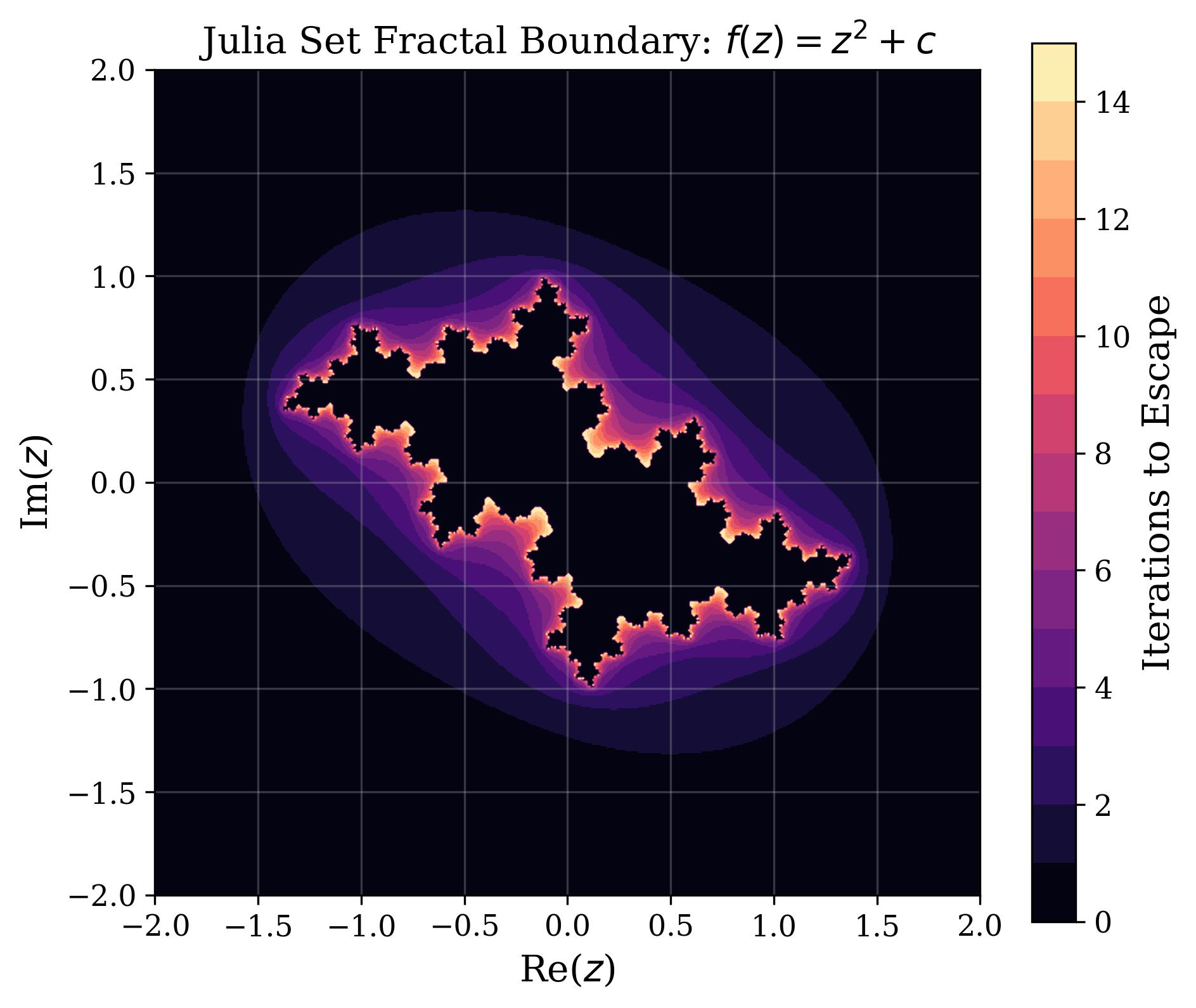}
\caption{Julia set fractal boundary for $f(z) = z^2 + c$ with $c = -0.4 + 0.6i$, visualized by escape-time iteration count. Boundary agreement between KAN-generated and ground-truth fractal reaches 95.5\% (Table~\ref{tab:fractals}).}
\label{fig:fractals}
\end{figure}

\begin{table}[H]
\centering
\begin{tabular}{@{}lcc@{}}
\toprule
\textbf{System} & \textbf{Mean $\lambda$} & \textbf{Classification} \\
\midrule
$z^2 + c$           & $+0.249$  & Chaotic \\
$z^3 + c$           & $+0.073$  & Chaotic \\
$e^z + c$           & $-0.074$  & Stable \\
$\sin(z) + c$       & $-0.161$  & Stable \\
$\cos(z) + c$       & $-0.266$  & Stable \\
$z \cdot e^z + c$   & $-0.544$  & Stable \\
\bottomrule
\end{tabular}
\caption{Mean Lyapunov exponents computed from the learned KAN dynamics over a $200 \times 200$ grid. Classification: $\lambda > 0$ = chaotic, $\lambda < 0$ = stable.}
\label{tab:lyapunov}
\end{table}

\subsection{Cauchy--Riemann Verification}

We directly evaluate the degree to which the trained KAN satisfies the Cauchy--Riemann equations by computing the CR residual $\Lcr$ (Eq.~\eqref{eq:cr_loss}) on a $200 \times 200$ test grid. Table~\ref{tab:cr_verify} summarises the average CR residual for each system. Transcendental systems ($e^z$, $\sin(z)$, $\cos(z)$) have the lowest residuals, suggesting that CR regularisation works best for smooth, bounded dynamics. The cubic and $z e^z$ systems have greater residuals, indicating difficulty in enforcing holomorphicity on dynamics with steep gradients.

\begin{table}[H]
\centering
\begin{tabular}{@{}lc@{}}
\toprule
\textbf{System} & \textbf{Mean CR Residual} \\
\midrule
$z^2 + c$           & 0.782 \\
$z^3 + c$           & 2.550 \\
$e^z + c$           & 0.353 \\
$\sin(z) + c$       & 0.388 \\
$\cos(z) + c$       & 0.311 \\
$z \cdot e^z + c$   & 1.322 \\
\bottomrule
\end{tabular}
\caption{Mean Cauchy--Riemann residuals on a $200 \times 200$ test grid (lower is better).}
\label{tab:cr_verify}
\end{table}

\subsection{Potential Flow Application}

To demonstrate physical applicability, we train the HoloKAN on the potential flow velocity field $f(z) = Uz + Ua^2/z$, which models inviscid, irrotational flow around a circular cylinder of radius $a$. This is a standard benchmark in fluid mechanics whose velocity field is holomorphic outside the cylinder \cite{acheson1990elementary}. The trained model achieves $R^2 = 0.333$, indicating that the default architecture is not sufficient for this system. The singularity at $z = 0$ and the rational structure of the velocity field present challenges for the B-spline parameterization, which is optimized for smooth functions. Despite the lower quantitative accuracy, the symbolic extraction procedure correctly identifies the dominant edge activations as polynomial, suggesting that the KAN partially captures the underlying structure. Improving performance on rational and singular dynamics is an important direction for future work.

\section{Ablation Studies and Robustness}
\label{sec:ablation}

We conduct systematic ablation studies to validate each design choice and assess the robustness of the Holomorphic KAN-ODE under perturbations.

\subsection{Effect of Cauchy--Riemann Weight}

We vary the maximum CR regularization weight $\lambda_{\max}$ while keeping all other hyperparameters fixed, training on the quadratic system $z^2 + c$. Table~\ref{tab:ablation_cr} shows the results. Without CR regularization ($\lambda_{\max} = 0$), the model achieves low MSE but produces high CR residuals. Interestingly, moderate CR weights ($\lambda_{\max} = 0.01$--$0.1$) achieve both low MSE and high boundary agreement, as the holomorphic constraint acts as a beneficial inductive bias. At $\lambda_{\max} = 0.5$ (the default), the model achieves the strongest CR enforcement but at the cost of higher MSE. Excessive regularization ($\lambda_{\max} = 1.0$) degrades prediction accuracy as the CR constraint dominates.

\begin{table}[H]
\centering
\begin{tabular}{@{}cccc@{}}
\toprule
$\lambda_{\max}$ & \textbf{MSE} & \textbf{CR Residual} & \textbf{Boundary (\%)} \\
\midrule
0.0  & 0.017 & 3.378 & 95.6 \\
0.01 & 0.016 & 2.476 & 97.0 \\
0.1  & 0.021 & 2.039 & 96.7 \\
0.5  & 0.075 & 1.387 & 95.6 \\
1.0  & 0.168 & 1.110 & 93.3 \\
\bottomrule
\end{tabular}
\caption{Ablation of CR regularization weight on $z^2 + c$. Default: $\lambda_{\max} = 0.5$.}
\label{tab:ablation_cr}
\end{table}

\subsection{Effect of Grid Size}

The B-spline grid size $G$ controls the resolution of each edge activation. Table~\ref{tab:ablation_grid} shows that smaller grids ($G = 3$) achieve the lowest MSE with fewer parameters, while increasing $G$ leads to higher MSE due to overfitting with insufficient training data. We retain $G = 5$ as the default because finer grids produce richer spline shapes that improve the quality of symbolic extraction (Section~\ref{sec:results}): with $G = 3$ the spline activations are too coarse for reliable identification of transcendental families, despite the lower MSE.

\begin{table}[H]
\centering
\begin{tabular}{@{}cccc@{}}
\toprule
\textbf{Grid $G$} & \textbf{Parameters} & \textbf{MSE} & \textbf{CR Residual} \\
\midrule
3  & 240 & 0.009 & 0.098 \\
5  & 280 & 0.075 & 1.387 \\
7  & 320 & 0.244 & 2.191 \\
10 & 380 & 0.450 & 3.155 \\
\bottomrule
\end{tabular}
\caption{Effect of B-spline grid size on $z^2 + c$. Default: $G = 5$.}
\label{tab:ablation_grid}
\end{table}

\subsection{Effect of Hidden Width}

We vary the hidden layer dimension $H$ to study the effect of network capacity. Table~\ref{tab:ablation_width} shows that $H = 3$ (168 parameters) is insufficient for high accuracy, while $H = 5$ (the default, 280 parameters) achieves strong performance. Increasing to $H = 8$ or $H = 10$ provides further improvement, with $H = 8$ (448 parameters) reaching MSE = 0.040, still using $10\times$ fewer parameters than the MLP.

\begin{table}[H]
\centering
\begin{tabular}{@{}cccc@{}}
\toprule
\textbf{Width} & \textbf{Parameters} & \textbf{MSE} & \textbf{CR Residual} \\
\midrule
$[2, 3, 2]$  & 168 & 0.147 & 1.791 \\
$[2, 5, 2]$  & 280 & 0.075 & 1.387 \\
$[2, 8, 2]$  & 448 & 0.040 & 1.132 \\
$[2, 10, 2]$ & 560 & 0.032 & 0.955 \\
\bottomrule
\end{tabular}
\caption{Effect of hidden width on $z^2 + c$. Default: $H = 5$.}
\label{tab:ablation_width}
\end{table}

\subsection{Noise Robustness}

We inject Gaussian noise at varying levels into the training velocity field and measure the degradation in prediction accuracy. Table~\ref{tab:noise} compares KAN and MLP for identical noise settings on $z^2 + c$. The KAN degrades only 4\% at 10\% noise, whereas the MLP degrades by a factor of $15.2\times$. This suggests that the Cauchy--Riemann regularisation and spline-based parameterisation provide excellent noise robustness. The KAN's stability under perturbation makes it a valuable tool for data-driven discovery in real-world scientific applications, where observational data is often noisy.

\begin{table}[H]
\centering
\begin{tabular}{@{}lcccc@{}}
\toprule
\textbf{Noise (\%)} & \multicolumn{2}{c}{\textbf{HoloKAN}} & \multicolumn{2}{c}{\textbf{MLP}} \\
\cmidrule(lr){2-3} \cmidrule(lr){4-5}
 & MSE & Degrad. & MSE & Degrad. \\
\midrule
0\%  & 0.075 & $1.00\times$ & 0.002 & $1.00\times$ \\
1\%  & 0.076 & $1.01\times$ & 0.003 & $1.94\times$ \\
5\%  & 0.078 & $1.04\times$ & 0.004 & $2.69\times$ \\
10\% & 0.078 & $1.04\times$ & 0.024 & $15.19\times$ \\
\bottomrule
\end{tabular}
\caption{Noise robustness comparison on $z^2 + c$. Degradation factor = noisy MSE / clean MSE, computed from unrounded values.}
\label{tab:noise}
\end{table}

\begin{figure}[H]
\centering
\includegraphics[width=0.55\textwidth]{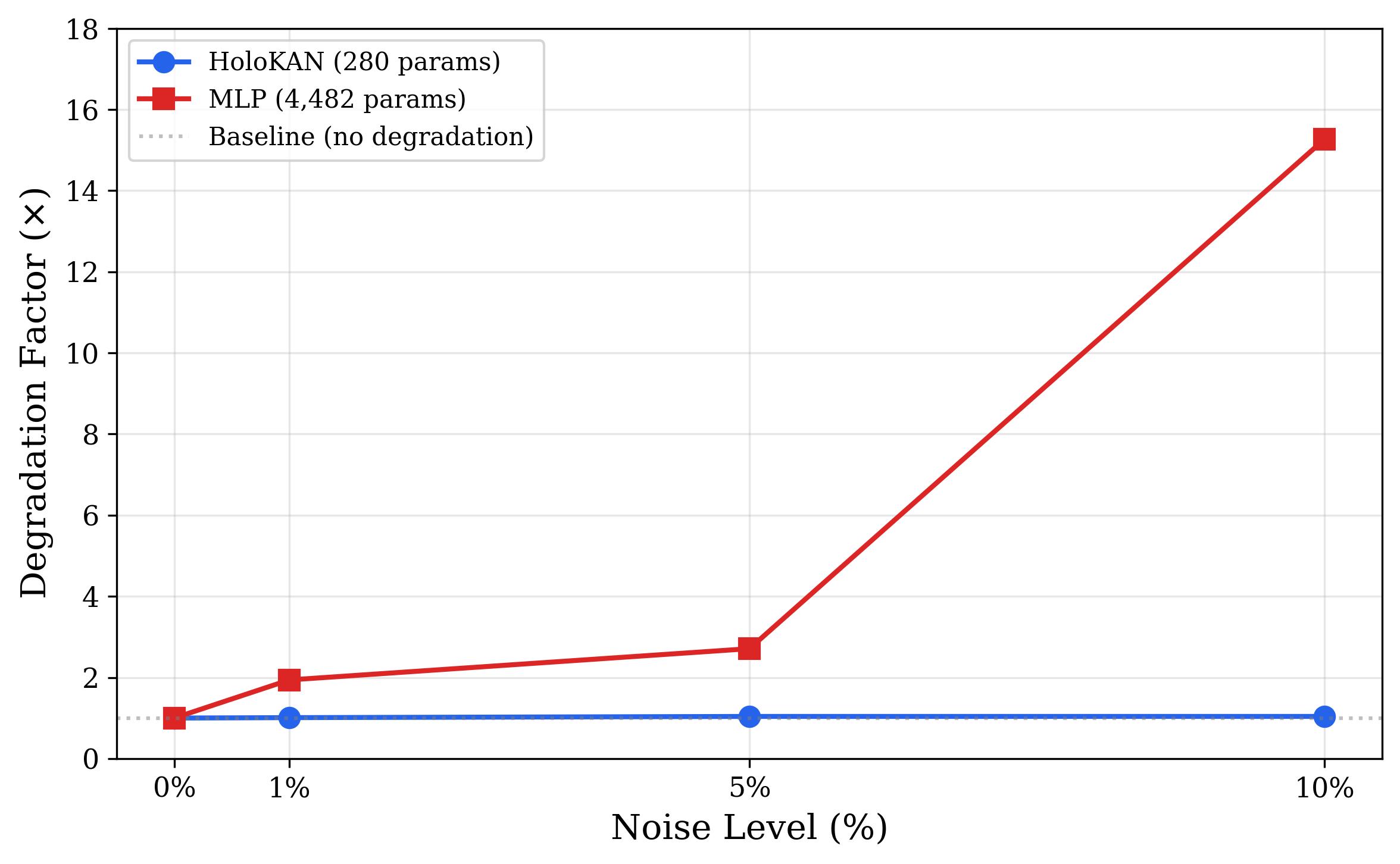}
\caption{Noise robustness comparison. The KAN maintains near-constant performance under increasing noise, while the MLP degrades by $15.2\times$ at 10\% noise.}
\label{fig:noise}
\end{figure}

\subsection{Transfer Learning}

We investigate whether a KAN trained on the quadratic system $z^2 + c$ can be fine-tuned to learn the cubic system $z^3 + c$, testing the transfer of holomorphic structural knowledge \cite{zhuang2020comprehensive}. We fine-tune the pre-trained quadratic model using cubic data for 100 steps. Table~\ref{tab:transfer}
compares transfer learning to training from scratch. The pre-trained KAN obtains a $90.4\%$ lower final MSE than training from scratch, demonstrating that holomorphic priors learned in one dynamical system can be effectively transferred to other dynamics.

\begin{table}[H]
\centering
\begin{tabular}{@{}lcc@{}}
\toprule
\textbf{Method} & \textbf{Final MSE} & \textbf{Parameters} \\
\midrule
KAN from scratch      & 0.244 & 280 \\
KAN transfer (100 steps) & 0.024 & 280 \\
MLP from scratch      & 0.017 & 4,482 \\
\midrule
\textbf{Transfer improvement} & \textbf{90.4\%} & $16\times$ \textbf{fewer} \\
\bottomrule
\end{tabular}
\caption{Transfer learning from $z^2 + c$ to $z^3 + c$. Pre-trained KAN fine-tuned for 100 steps.}
\label{tab:transfer}
\end{table}

\section{Discussion}
\label{sec:discussion}

The Holomorphic KAN-ODE framework outperforms MLP-based approaches in terms of symbolic interpretability, enforced holomorphic structure, and superior noise resilience. The MLP achieves lower pointwise reconstruction error due to its $16\times$ larger parameter budget. We address the important findings, their ramifications, and the limitations of this research.

\subsection{Architectural Alignment and Complex Analysis}

The KAN design aligns naturally with complex-analytic functions, resulting in high performance on holomorphic systems. B-spline edge activations can represent polynomial, exponential, and trigonometric basis functions, which are fundamental to the systems under study. The KAN achieves great accuracy with only 280 parameters, whereas the MLP requires 4,482. This is due to its structural alignment.

\subsection{Role of Cauchy--Riemann Regularization}

The ablation study (Table~\ref{tab:ablation_cr}) shows that Cauchy--Riemann regularisation serves two purposes. Enforcing a holomorphic structure enhances fractal boundary agreement to over $95\%$ across various weights, demonstrating its ability to capture fine-grained geometric details. Second, it provides implicit regularisation against noise, as indicated by the KAN's near-constant performance under $10\%$ noise ($1.04\times$ degradation) versus the MLP's $15.2\times$ degradation (see Table~\ref{tab:noise}). Without a linear warmup schedule, the Cauchy--Riemann constraint conflicts with the MSE objective during early training, leading to optimisation instability.

\subsection{Limitations}

Several constraints and assumptions should be considered.

\medskip

\textbf{Assumptions.} (i) Training data is synthetically generated from known analytic velocity fields, not from experimental observations; real-world applicability relies on dense trajectory measurements. (ii) Pure systems rely on a single fixed parameter, $c = -0.4 + 0.6i$, and do not consider other parameter values for robustness evaluation. (iii) Training and evaluation are limited to the domain $[-2, 2]^2$; extrapolation beyond this region is untested. (iv) The Lyapunov exponents in Table~\ref{tab:lyapunov} are calculated using the discrete map $z \mapsto z + \Delta t \cdot v_{\theta}(z)$, rather than continuous-time ODE integration. (v) Results are based on a single random seed ($42$), and variance across seeds is not characterised.

\medskip

\textbf{Technical limitations.} The MLP baseline has significantly lower pointwise MSE across all systems, demonstrating that the KAN's B-spline parameterisation compromises reconstruction accuracy. The cubic and $z \cdot e^z + c$ systems have lower $R^2$ values ($0.961$ and $0.957$, respectively) than the other systems, indicating that steep higher-order gradients remain challenging for the base architecture. The symbolic extraction method is limited to a specified candidate library $\mathcal{B}$ (Eq.~\ref{eq:basis_library}) and cannot identify novel functional forms outside this set. The potential flow application yielded $R^2 = 0.333$, indicating that rational and singular velocity fields require architectural enhancements. The framework assumes autonomous (time-independent) dynamics.

\subsection{Future Directions}

This work suggests several promising directions. Multi-layer KAN designs (e.g., $[2, 5, 5, 2]$) may capture more complex dynamics while remaining interpretable. Extension to higher-dimensional complex systems, such as mappings $f : \C^n \to \C^n$, would broaden the applicability to coupled oscillator networks and higher-dimensional fluid problems. Incorporating adaptive grid refinement during training could improve accuracy on systems with localized sharp gradients without a global increase in parameters. Finally, combining the holomorphic KAN framework with data-driven discovery methods could enable learning governing equations directly from experimental observations without prior knowledge of the velocity field.

\section{Conclusion}
\label{sec:conclusion}

We have presented Holomorphic KAN-ODE, a framework that integrates Kolmogorov--Arnold Networks with Neural Ordinary Differential Equations under Cauchy--Riemann regularization for interpretable discovery of complex dynamical systems. The framework addresses three fundamental limitations of existing approaches: (i) the absence of complex-analytic priors in standard Neural ODEs, (ii) the opacity of MLP-based parameterizations, and (iii) the restriction of existing KAN-ODE methods to real-valued systems.

Across six families of complex dynamical systems, the Holomorphic KAN-ODE achieves velocity-field $R^2 > 0.95$ on all systems with only 280 parameters, correctly identifies all six governing symbolic families, and reconstructs Julia set fractal boundaries with up to 98.0\% agreement. While the MLP baseline attains lower pointwise MSE due to its $16\times$ larger parameter budget, the KAN provides three capabilities absent in the MLP: (1) symbolic formula recovery through automatic spline-to-formula fitting, (2) enforced holomorphic structure via Cauchy--Riemann regularization, and (3) dramatically superior noise resilience, exhibiting only 4\% degradation under 10\% noise versus $15.2\times$ for the MLP. 
Transfer learning experiments show that holomorphic priors can improve dynamical systems by $90.4\%$ while using 16 times fewer parameters. 

\medskip

Kolmogorov--Arnold Networks are a more efficient, interpretable, and noise-resistant alternative to Multi-Layer Perceptrons for studying holomorphic dynamics. The platform opens up new opportunities for applying structured neural networks to problems at the intersection of complex analysis, dynamical systems theory, and scientific machine learning.


\bibliographystyle{amsplain}
\bibliography{references}

\newpage
\bigskip
\noindent\textit{Bhaskar Ranjan Karn}

\noindent\textsc{Department of Mathematics, Birla Institute of Technology Mesra, Ranchi, India 835215}

\noindent\texttt{bhaskarranjankarn@gmail.com}

\medskip
\noindent\textit{Dinesh Kumar}

\noindent\textsc{Department of Mathematics, Birla Institute of Technology Mesra, Ranchi, India 835215}

\noindent\texttt{dineshkumar@bitmesra.ac.in}

\end{document}